%% file: main.tex
\documentclass[runningheads]{llncs}
\usepackage[T1]{fontenc}
\usepackage{graphicx}
\usepackage{subcaption}
\usepackage{booktabs}
\usepackage{amsmath}
\usepackage{float}
\usepackage{pxfonts}
\usepackage{algorithm}
\usepackage{algorithmic}
\usepackage{multirow}
\usepackage{arydshln}
\usepackage[table,xcdraw]{xcolor}
\usepackage{changes}
\usepackage{hyperref}
\usepackage{xr-hyper}
\usepackage[misc]{ifsym}

\newcommand{\R}{\ensuremath{\mathbb{R}}}
\floatname{algorithm}{Pseudo-code}
\newenvironment{AlgoBodySmall}{%
  \begingroup
  \fontsize{8.5pt}{10.2pt}\selectfont
}{%
  \endgroup
}

\definechangesauthor[name={Milad}, color=orange]{milad}
\definechangesauthor[name={Guillaume}, color = purple]{guillaume}

\makeatletter
\newcommand{\@chapapp}{\relax}%
\makeatother

\usepackage{mwe}

\begin{document}
\title{A Machine Learning Framework for Turbofan Health Estimation via Inverse Problem Formulation}

\titlerunning{Turbofan Health Estimation using Machine Learning}


\author{
  Milad Leyli-Abadi\inst{1}\thanks{Corresponding author} \and
  Lucas Thil\inst{1,2} \and
  S\'ebastien Razakarivony\inst{3} \and
  Guillaume Doquet\inst{3} \and
  Jesse Read\inst{2}
}

\authorrunning{Leyli-Abadi et al.}

\institute{
  Institut de recherche technologique SystemX, Palaiseau, France\\
  \email{milad.leyli-abadi@irt-systemx.fr}
  \and
  \'Ecole Polytechnique, LIX, Palaiseau, France\\
  \email{thil@lix.polytechnique.fr}
  \and
  Safran Tech, Magny-les-Hameaux, France\\
  \email{sebastien.razakarivony@safrangroup.com}
}






\maketitle              

\begin{abstract}

Estimating the health state of turbofan engines is a challenging ill-posed inverse problem, hindered by sparse sensing and complex nonlinear thermodynamics. Research in this area remains fragmented, with comparisons limited by the use of unrealistic datasets and insufficient exploration of exploitation of temporal information. This work investigates how to recover component‑level health indicators from operational sensor data under realistic degradation and maintenance patterns. To support this study, we introduce a new dataset that incorporates industry-oriented complexities such as maintenance events and usage changes. Using this dataset, we establish an initial benchmark that compares steady‑state and nonstationary data‑driven models, and Bayesian filters --classic families of methods used to solve this problem. In addition to this benchmark, we introduce self-supervised learning (SSL) approaches that learn latent representations without access to true health labels, a scenario reflective of real-world operational constraints. By comparing the downstream estimation performance of these unsupervised representations against the direct prediction baselines, we establish a practical lower bound on the difficulty to solve this inverse problem. Our results reveal that traditional filters remain strong baselines, while SSL methods reveal the intrinsic complexity of health estimation and highlight the need for more advanced and interpretable inference strategies. For reproducibility, both the generated dataset and the implementation used in this work are made accessible.\footnote{\url{https://sandbox.zenodo.org/records/469530}}\footnote{\url{https://github.com/ConfAnonymousAccount/ECML_PKDD_2026_TurboFan}}.

\keywords{Turbofan engine health monitoring \and Realistic simulation data \and Time series modeling \and Representation learning \and State-space models}
\end{abstract}

\section{Introduction}

Monitoring the health of turbomachines is critical for ensuring reliable aircraft engine monitoring and enabling predictive maintenance. However, modern aircraft operate in mechanical, thermal, and weight constrained environments, which severely limit the number and placement of onboard sensors. As a consequence, available measurements provide only a partial view of the engine's internal state, making health estimation an ill-posed and under-determined problem: several distinct health states may yield nearly identical sensor values. 

Engine health is commonly characterized by a thermodynamic model that relates the different measured temperatures and pressures in and out of the engine, depending on the efficiencies and mass flows of each component of the engine (compressors, turbines). Variations in these two quantities serve as \emph{health indicators}, reflecting the level of degradation in each of these components. Health monitoring therefore needs to solve a challenging inverse problem : retrieving the efficiencies and mass flows from a sparse set of measurements.

While the underlying thermodynamics of a turbofan engine are well understood, solving this inverse problem in practice remains challenging. Traditionally, physics-based methods like Gas Path Analysis (GPA) \cite{Urban1969} or Kalman filters \cite{liu2022aero} heavily rely on accurate prior assumptions and can struggle with modeling errors. In contrast, purely data-driven approaches \cite{de2022intelligent} offer flexibility by learning complex representations from data, but often fail to capture domain-specific degradation mechanisms without vast amounts of realistic time-series.  


A critical barrier to advancing this field is the lack of a public benchmark that captures the complex degradation patterns found in real-world turbomachinery. Existing datasets often lack the necessary fidelity to validate algorithms, primarily because they do not provide verifiable ground-truth health indicators for the inverse problem. Consequently, it remains difficult to compare different methodological practices and gauge progress towards practical, deployable solutions. In this paper, we study the inverse problem of health estimation of turbofan components from sparse sensor data. Our main contributions are:

\begin{itemize}
    \item \textbf{A realistic turbofan dataset:} We release a public dataset reflecting the challenges of real-world engine health monitoring with realistic degradation mechanisms. 
    \item \textbf{Comprehensive baseline evaluation:} Conduct a thorough evaluation of established methods applied to the inverse problem of direct estimation the ground health states from sparse measurements. 
    \item \textbf{Self-Supervised Learning:} We investigate how effectively representations learned on sensor measurement alone can effectively recover the ground truth health states.
\end{itemize}

The remainder of the article is organized as follows: the related work is presented in Section \ref{sec:related_work}. The dataset and the generation process are described in Section \ref{sec:data_and_scenario_design}. The methodologies analyzed and experimented in this work are presented in Section \ref{sec:models}. The experimental settings and results are shown in Section \ref{sec:experiments}. In Section \ref{sec:conclusion}, we conclude the paper giving insights for future works.

\section{Related Works}
\label{sec:related_work}
The turbofan engine performance inverse problem (TEP-IP) has been addressed for more than half a century (\cite{Urban1969}), giving rise to several families of methods, from physics‑based filtering to machine‑learning regression and hybrid formulations. We first review the main approaches and then broaden the scope with a peek on health-state embeddings and datasets of engine health monitoring. For more detailed information, we redirect the readers to recent surveys \cite{Vu2025,Soleimani2025}.

\subsection{Embedded model-based approaches}
The most common works on solving the TEP-IP is the use of bayesian estimation, and more particularly Kalman Filters (KF). KF models the health state as a latent state evolving through a process model $f$ and observed through an observation model $h$, using the formalism of the state space model:
\begin{equation}
    x_{t+1} = f(x_t) + v,     y_{t} = h(x_t) + w.
\end{equation}
 In our case, $x_t$ represents the component efficiencies and mass flows, while $y_t$ corresponds to sensor measurements. $v$ and $w$ are random variables. The observation model $h$ is usually derived from the thermodynamic simulator, whereas the system‑level degradation model $f$ requires assumptions or simplifications, that are not always easy to define (for complete notations, see~\ref{app:notations}). Many papers using many variants of KF have been published over the years (\cite{simon2008comparison,lu2016improved,liu2022aero}, making the Kalman Filter the strongest baseline on this matter. However, these papers do not provide the datasets to reproduce their experiments.

\subsection{Machine learning approaches}
An alternative line of work uses supervised learning to approximate the inverse of the thermodynamic model (\cite{castillo2021data,loboda2011polynomials,vu2024aircraft}). These methods use the thermodynamic model by first building a dataset of pairs ($x, y$) and then learn $g$ such that $x = g(y)$. When the thermodynamic model is non‑invertible, the behavior of the estimator depends strongly on architectural choices and regularization strategies. Some more recent works tries to adapt estimation strategies, in particular with Reinforcement Learning \cite{tian2022real,schirru2025adaptive}, mixing the classic KF and data-driven approaches. However, all these methods have a low explainability.

\subsection{Embedding health states and self-supervised learning} 
Beyond direct inversion, much research targets Remaining Useful Life (RUL) prediction but often neglects explainability of intermediate states. To represent current condition, others build data-driven health indicators \cite{Datong_Liu_2015}; however, since these are not a core requirement, they lack inherent physical meaning and require post-hoc explanation, limiting diagnostic value. Similarly, some works embed engine states data-drivenly (\cite{De_Pater_Mitici_2022,Pillai_Vadakkepat_2021}), but without links to physics-based models, these embeddings also depend on external interpretation and cannot serve as directly meaningful health indicators.

Further works tend to construct features such as reconstruction error or novelty detection to act as a proxy for a health indicator extracted from the output or latent space \cite{Costa_Sanchez_2022,Thil_Read_Kaddah_Doquet_2026}, or induce a temporal aspect in the latent representations \cite{Lee_Park_Lee_2024} by computing soft assignment pairs in the observation space. Other methods such as the JEPA architecture focus in predicting directly in the latent \cite{Bordes_Garrido_Kao_Williams_Rabbat_Dupoux_2025,Assel_Ibrahim_Biancalani_Regev_Balestriero_2025}, promising better representation and planning abilities \cite{Assran_Bardes}, components though after by industrials. 


\subsection{Datasets}

Most reproducible studies rely on the NASA C‑MAPSS dataset \cite{Saxena2008}, which provides run‑to‑failure trajectories generated from a thermodynamic simulator but assumes a non-realistic and specific degradation pattern. Furthermore, the underlying engine model has some access restrictions. The more recent New‑C‑MAPSS dataset \cite{Chao2021} introduces realistic flight profiles and multiple failure modes, though still with fast degradation dynamics and no maintenance interventions, as any normal engine would encounter in its life. Finally, the PHM2025 challenge dataset \cite{PHM2025Challenge} provides more accurate degradation, with simulation‑based maintenance events. However, it presents only a handful of examples and fixed degradation rates, making it less suited for learning richly parameterized health dynamics.

\section{Data and scenario design}
\label{sec:data_and_scenario_design}

\subsection{Turbomachine model}

Our experiments rely on the OpenDeckSMR simulator \cite{Psaropoulos_OpenDeckSMR_2025}, a steady‑state turbofan performance model designed following industrial engine‑simulation practices. This tool solves the thermodynamic balance equations across all engine components to compute temperatures, pressures, power and rotation speeds. Given a set of health parameters—component efficiencies and mass flows—and a specified operating condition (e.g., take‑off, climb, cruise), the simulator provides the corresponding sensor measurements. This makes it suitable for generating consistent synthetic data for health‑state estimation experiments. The simulator uses ten Health Indicators (HIs), five pairs of (efficiency, mass flow) coefficients of degradation.


\subsection{Data generation}
\begin{figure}
    \centering
    \includegraphics[width=0.9\linewidth]{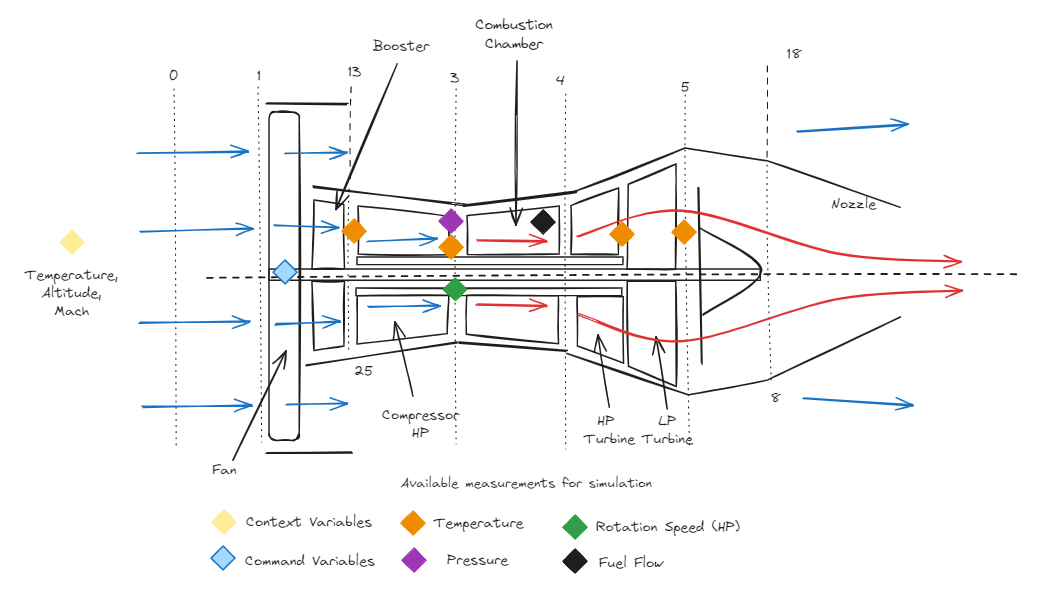}
    \caption{Turbofan engine; with the courtesy of the team of OpenDeckSMR.}
    \label{fig:turbofan}
\end{figure}

Given a health state at timestep $\boldsymbol{x}_t \in \R^{10}$ and an Operational Condition $\boldsymbol{OC}$, the physical simulator described in the previous section allows to simulate the sensor values $\boldsymbol{y}_t = h(\boldsymbol{x}_t, \boldsymbol{OC})$ at different locations of the engine. This steady-state computation scheme is not adapted to model the real-world problem. Generally, the turbofan engines may degrade gradually over time based on different conditions, usage and maintenance operations performed during their lifetime. The degradation speed may also change with respect to these conditions and regarding the different components in the engine. For example the high pressure compressor (Compressor HP in Figure \ref{fig:turbofan}) degrades faster than the other components, as it is exposed to much higher temperature.

\begin{algorithm}
\caption{Data generation process}
\begin{AlgoBodySmall}
\begin{algorithmic}[1]

\STATE \textbf{Input:} \# sequences ($N$), \# timesteps  ($T$), Operational condition ($C$)
\STATE \textbf{Output:} Degradation trajectories (health indicators) $\boldsymbol{X}\in\R^{10}$\\\hspace{1.15cm} Simulated sensor values (measures) $\boldsymbol{Z}\in\R^{7}$
\STATE Select an operational condition in the set \{\textit{Cruise}, \textit{Takeoff}, \textit{Climb1}, \textit{Climb2}\}
\STATE Initialize the health indicators $\boldsymbol{X}_0 \gets \boldsymbol{0}.$ (Full health state)
\REPEAT
\FOR{each timestep $t = 1..T$}
    \FOR{each health indicator $x$ in $\boldsymbol{x}$}
        \IF{$t$ matches a maintenance event}
            \STATE $x_t \gets$ recover a random fraction $\lambda \in [0.6, 0.8]$ of past degradation and update
        \ENDIF
        \IF{speed change frequency reached}
            \STATE $\mu, \sigma$ $\gets$ resample degradation speed with state-specific probability distribution
        \ENDIF
        \STATE $m$, $\epsilon$ $\gets$ Sample slope $\mathcal{N}(\mu, \sigma)$ and a Gaussian noise wrt. the selected speed
        \STATE $x_t \gets$ $m + \epsilon$  Update and clip health indicator {\scriptsize(using boundary values in \ref{app:sim_parameters})}
        \STATE $z_t \gets \Psi(x_t, C)$ Compute the sensor values using the simulator
        \STATE $\tilde{z}_t \gets z_t + \epsilon$ Add a bounded noise into each sensor channel
    \ENDFOR
\ENDFOR
\UNTIL required number of sequences $N$
\RETURN All trajectory sequence $\boldsymbol{X}$ and corresponding simulated sensor values $\boldsymbol{Z}$
\end{algorithmic}
\end{AlgoBodySmall}
\label{algo:data_gen}
\end{algorithm}

To be representative of realistic problem configuration, in this paper, we design degradation trajectories scenarios taking into account the above mentioned considerations. In this sense, each degradation trajectory $S$ represents a multivariate time series of dimension $10$ (number of health indicators corresponding to different components of the engine). Each component of the engine has its own specific degradation pattern and boundaries (minimum and maximum authorized values). Three different degradation speeds (i.e., \textit{slow}, \textit{normal} and \textit{fast}) are considered. The components may degrade following a probability distribution over the three speed values and also to transition from one speed to another with a specific frequency (e.g., every 100 timesteps). Maintenance operations could also take place after certain time steps, picking a random value within the interval $[200;500]$. The maintenance events allow to partially recover the previous health state, which is controlled by a coefficient selected randomly in the interval $[0.6;0.8]$. The generation procedure is summarized in pseudo-code~\ref{algo:data_gen} and the generated data can be obtained through this \href{https://sandbox.zenodo.org/records/469530}{link}~\cite{anonymous_dataset_2026}.

\begin{figure}[H]
    \centering
    \includegraphics[width=1\linewidth]{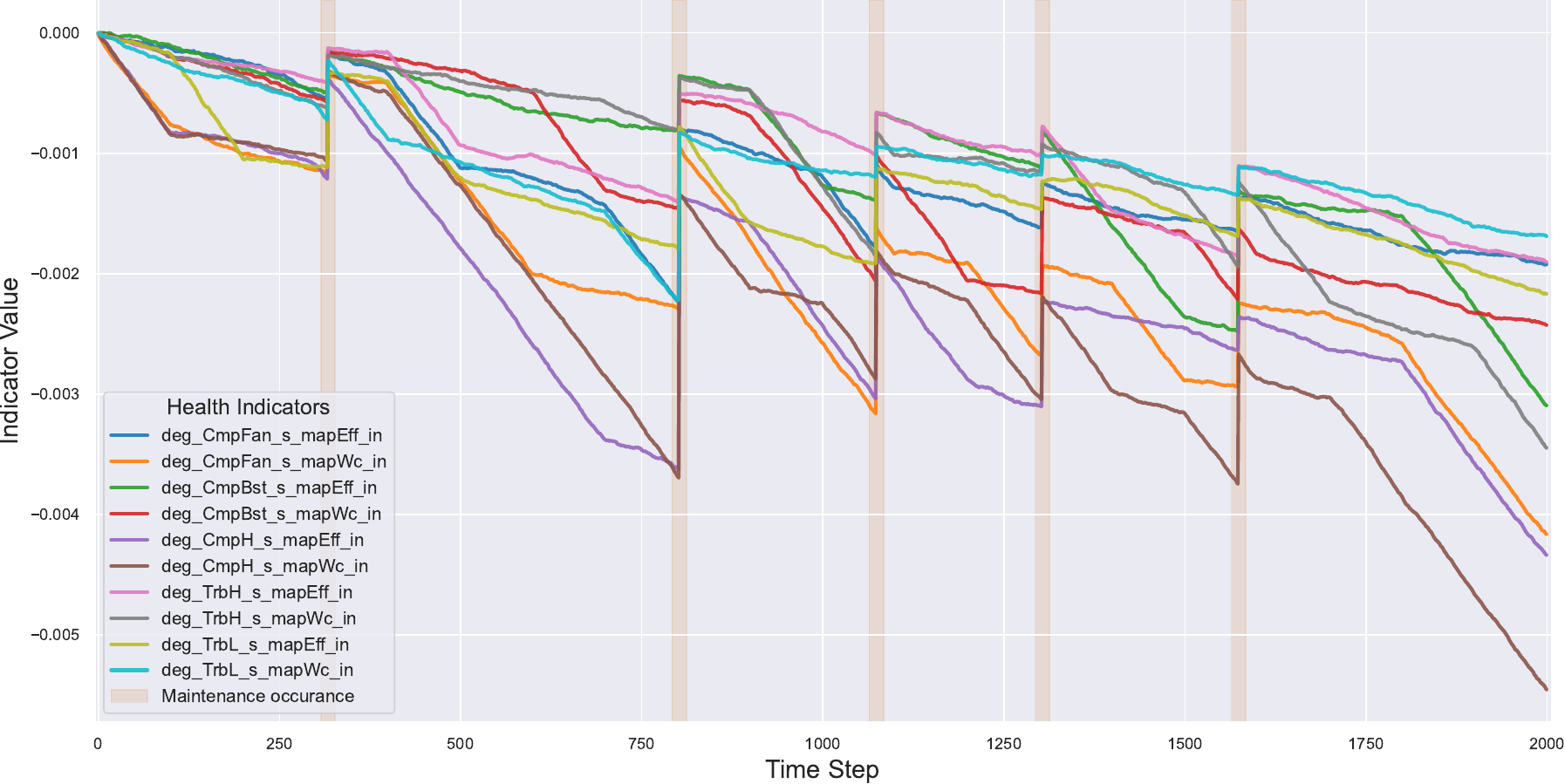}
    \caption{Generated degradation trajectories for 10 health indicators and 1 engine}
    \label{fig:trajectories}
\end{figure}

For the sake of our experiments and based on the above-mentioned parameters, an ensemble of more than 500 trajectories of health indicators are generated with the maximum length of 2000 time steps, on four operating conditions. An example of a trajectory is shown in Figure \ref{fig:trajectories}. As can be seen, the maintenance events could occur during the lifetime of the engine (the vertical boxes), where we observe that the engine recovers a percentage of its health state from previous operation period. It should be noted that some of the trajectories may have different (shorter) lengths, when one of the health indicators violates the authorized boundary values. 

\begin{figure}
    \centering
    \includegraphics[width=1\linewidth]{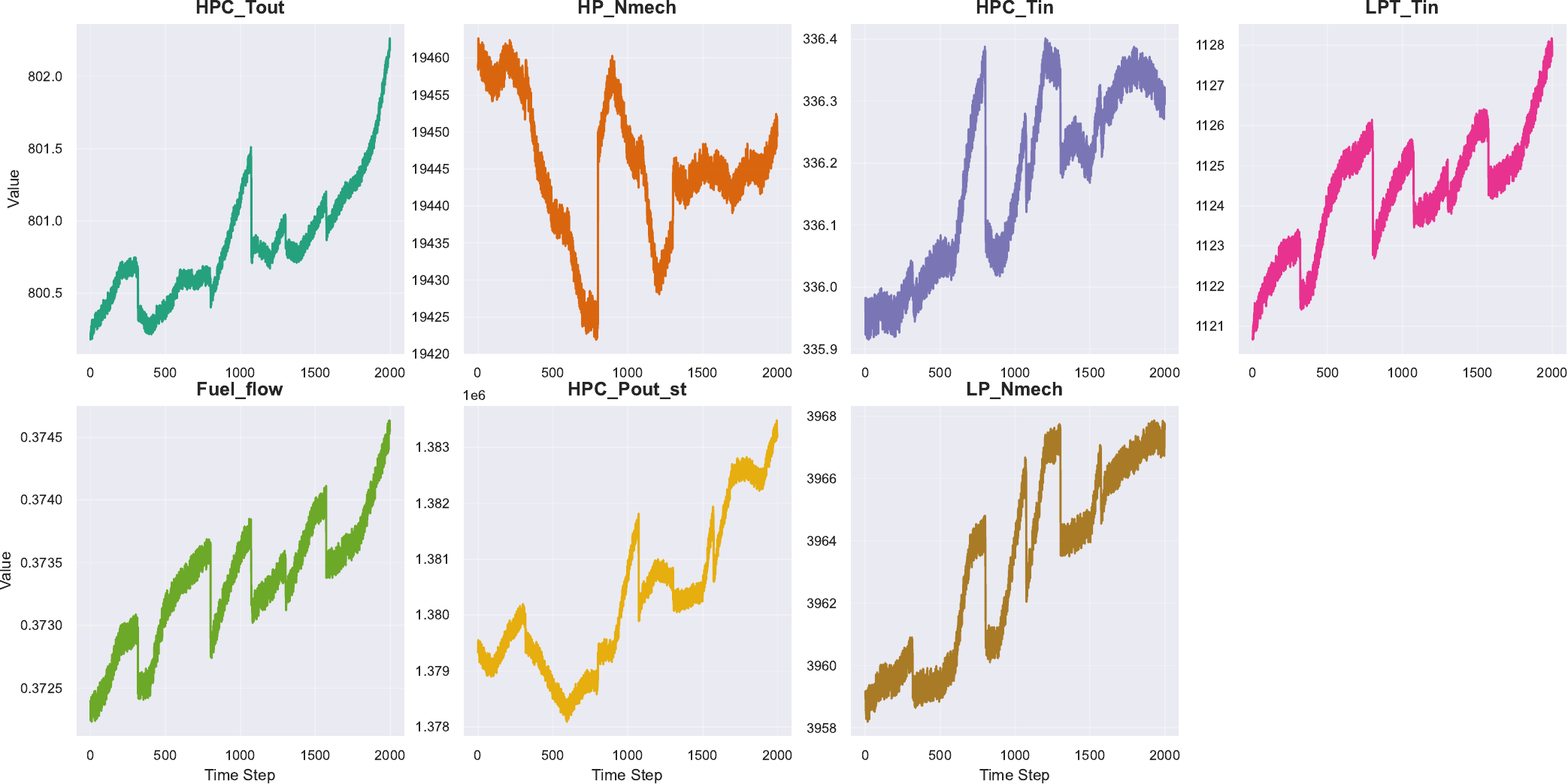}
    \caption{Simulated sensor values (measurements) for the trajectory in Figure~\ref{fig:trajectories} and corresponding to \textit{Cruise} operational condition.}
    \label{fig:sensor_measurements}
\end{figure}

For each trajectory, the sensor measurements are simulated using Turbomachine model and including 4 operational conditions, i.e., \textit{Cruise}, \textit{Takeoff}, \textit{Climb1}, \textit{Climb2}. The measurements for \textit{Cruise} are shown in Figure \ref{fig:sensor_measurements}. As can be seen, we added a bounded noise $\epsilon$ to each sensor measurement $y$ to approach the real-world condition as follows:
\begin{equation}
    \Delta = y_{max} - y_{min};
    \qquad
    \tilde{y}(t) = y(t) + \epsilon(t),
    \qquad
    \epsilon(t) \sim \mathcal{U}\left(-\gamma \Delta, \gamma \Delta\right),
\end{equation}
where the $\mathcal{U}$ represents the uniform distribution and gamma controls the noise level which is set to $0.02$. 

\begin{figure}
    \centering
    \includegraphics[width=.9\linewidth]{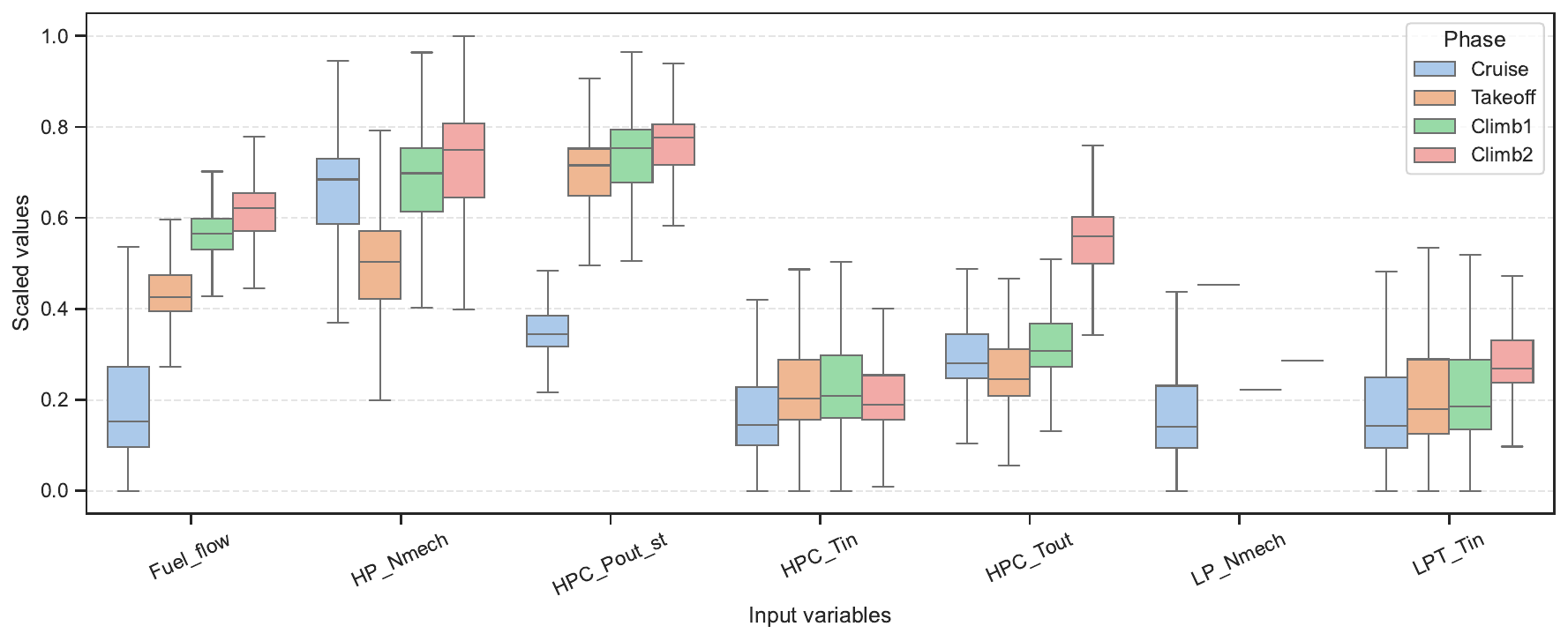}
    \caption{Sensor variables (measurements) distribution grouped by 4 flight phases.}
    \label{fig:sensor_dist}
\end{figure}

The distribution of the seven sensor variables and four operational conditions is shown in Figure \ref{fig:sensor_dist}. For the sake of comparison, these features are scaled using Min-Max normalization. In this graphic, each sensor exhibits a distinct distribution and variability range, reflecting the different physical quantities they measure within the turbofan system. Some sensors, such as \textit{LPT\_Tin} and \textit{HP\_Nmech}, display relatively narrow interquartile ranges, indicating stable behavior with limited fluctuation across trajectories. In contrast, variables like \textit{HP\_Tout}, \textit{Fuel\_flow}, and \textit{HPC\_Pout\_st} show broader spreads, suggesting higher variability driven by operating conditions or engine load changes. The presence of asymmetric boxes and extended whiskers in several sensors highlights moderate skewness or occasional extreme values. The distribution of health indicators is shown in the Appendix Figure~\ref{fig:hi_dist}.

\section{Models}
\label{sec:models}
\subsection{Baselines}

\subsubsection{Steady-state approaches}
\label{sec: steady-state}
The steady-state hypothesis considers no temporal dependence between the observations as can be seen in Figure~\ref{fig:steady_state} and each sample is independent and identically distributed (i.i.d). In this configuration, we infer directly the HI state ($x_i$) from the sensor measurements ($y_i$). In our experiments, we have considered Gradient Boosting (GB), an ensemble and decision tree-based approach, and also a Multi-Layer Perceptron (MLP) which is a deep neural network-based approach. 

\subsubsection{Non-stationary approach}
\label{sec: non-stationary}
\begin{figure}
     \centering
     \begin{subfigure}{0.48\textwidth}
         \centering
         \includegraphics[width=\linewidth]{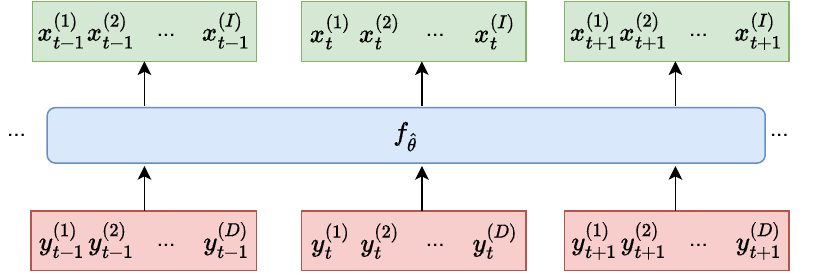}
         \caption{Steady-state hypothesis}
         \label{fig:steady_state}
     \end{subfigure}
     \begin{subfigure}{0.48\textwidth}
         \centering
         \includegraphics[width=\linewidth]{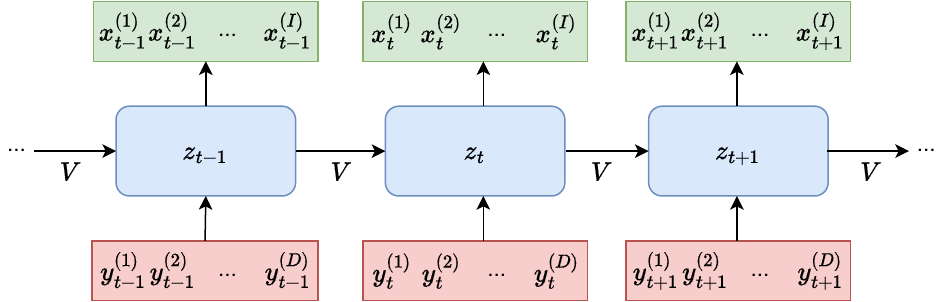}
         \caption{Non-stationary data hypothesis}
         \label{fig:temporal_based}
     \end{subfigure}
     \caption{Two model categories considering stationary and non-stationary hypotheses. Based on steady-state hypothesis, the model finds a mapping between inputs and outputs. Based on non-stationary hypothesis, the model takes into account the temporal dynamics over time for prediction of outputs.}
     \label{fig:partial_information_models}
\end{figure}

In contrast to the steady-state assumption, temporal-based approaches explicitly model the temporal dependencies between successive observations, thereby adhering to a non-stationary hypothesis. As illustrated in Figure~\ref{fig:temporal_based}, these models assume that the current health indicator (HI) state ($x_t$) cannot be inferred solely from the instantaneous measurement ($y_t$), but rather depends on the historical sequence of sensor readings $(y_{t-L+1}, \ldots, y_t)$ and their underlying temporal dynamics. This formulation enables the model to capture trends, degradation patterns, and short or long-term dependencies that are essential in prognostics scenarios where system behavior evolves over time with the presence of maintenance events. To exploit this temporal structure, we consider recurrent or sequence-based architectures capable of processing ordered data. In our experiments, we utilize Gated Recurrent Units (GRU), which is a widely adopted recurrent neural network (RNN) variant designed to mitigate vanishing gradient issues and effectively retain relevant information over long horizons. 

\subsubsection{Bayesian Filtering}
\label{sec: kalman}
In this section, we opt for Kalman Filter method, and more specifically we use an Unscented Kalman Filter (UKF). UKF is a non-linear version of UKF, well suited for our problematic, where the simulator is not linear.
We followed the standard formulation and squared-root implementation of \cite{van2001square}. The system is defined as:
\begin{equation}
    x_{t} = g(x_{t-1}) + w_{t}, \qquad
    y_{t} = h(x_{t}) + v_{t},
\end{equation}
where $x_t$ denotes the engine health state, $y_t$ the observed sensor measurements, and $w_t \sim \mathcal{N}(0,Q)$, $v_t \sim \mathcal{N}(0,R)$.  
The observation function $h$ corresponds to the OpenDeckSMR simulator, while the state-transition function $g$ is taken as the identity, a common assumption when no explicit degradation dynamics are available. 
The process noise covariance $Q$ is set to a diagonal matrix with value $10^{-7}$, and $R$ is derived from the known sensor noise specifications, divided by 10. These $Q$ and $R$ values where obtained by shallow grid-search. UKF hyperparameters follow the canonical settings typically used for mildly nonlinear systems ($(\alpha=1, \beta=2, \kappa=0)$).
\footnote{We do not address the simulation-to-reality gap (i.e., inaccuracies in $R$, $Q$, or the models $g$ and $h$). Even in this favorable configuration, our results show that the UKF still leaves room for improvement, making this an interesting direction for future research.}




\subsection{SSL Methods}
\begin{figure}[ht]
    \centering
    \includegraphics[width=\textwidth]{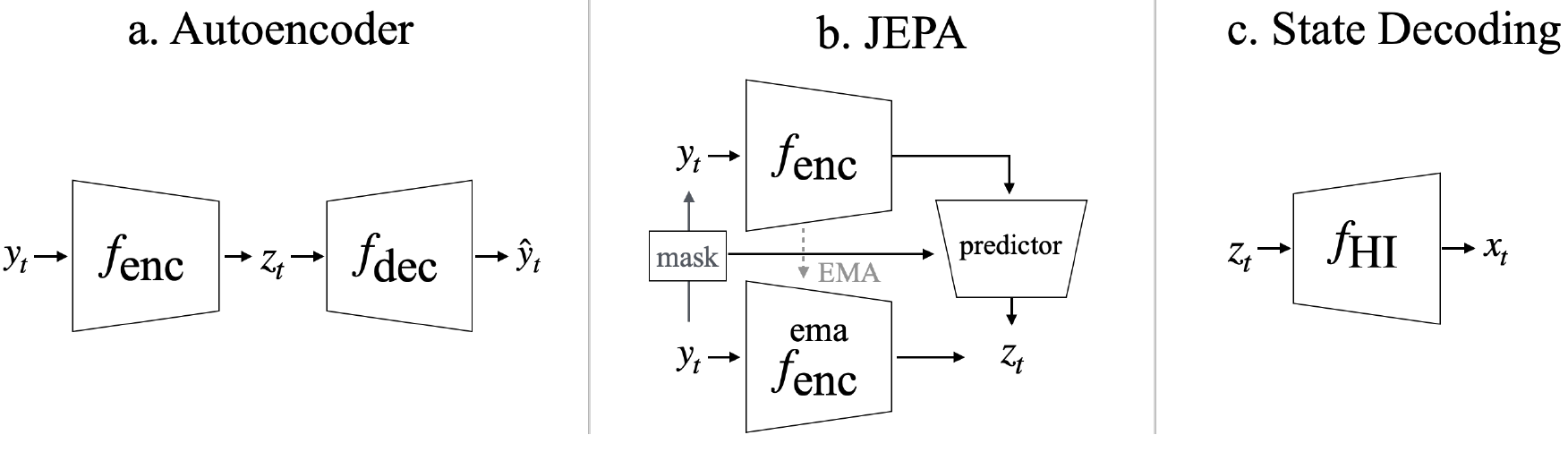}
    \caption{\textbf{SSL Approaches.} 
    \textbf{a. Autoencoder:} trained with reconstruction loss on $y_t$ to learn latent $z_t$. 
    \textbf{b. JEPA:} predicts masked patches in latent space to learn $z_t$. 
    \textbf{c. State Decoding:} predicts $x_t$ from frozen $z_t$ to evaluate representations.
}
    \label{fig:ssl_approach}
\end{figure}

In many industrial scenarios, access to ground-truth health indicators is unavailable, requiring models to learn representations from sensor data alone. We employ two self-supervised learning approaches: an autoencoder that operates in the observation space (reconstruction) and a JEPA architecture that predicts representations in the latent space. The extracted latents \(z_t\) are used for the downstream HI estimation task, as illustrated in Figure~\ref{fig:ssl_approach}.

\subsubsection{Autoencoder (AE)}
learns a compressed representation \(z_t\) of the input sensor data \(y_t\) by training to reconstruct the original input. The encoder \(f_{\text{enc}}\) maps \(y_t\) to a latent representation \(z_t = f_{\text{enc}}(y_t)\), and the decoder \(f_{\text{dec}}\) reconstructs the input as \(\hat{y}_t = f_{\text{dec}}(z_t)\). The training objective minimizes the reconstruction error:
\[
\mathcal{L}_{\text{AE}} = \|y_t - \hat{y}_t\|_2^2
\]
This loss operates entirely in the observation space without requiring ground-truth labels \(x_t\). 

\subsubsection{Joint Embedding Predictive Architecture (JEPA)}
While the autoencoder reconstructs in the observation space, JEPA \cite{Assel_Ibrahim_Biancalani_Regev_Balestriero_2025} learns by predicting the embeddings of masked patches in a latent space. This approach is well-suited for temporal sensor data, where relationships between time steps can be exploited. We adopt a Video JEPA (VJEPA) variant \cite{bardes2023v,Assran_Bardes} 
adapted for multivariate time series with multiple sensors and operating contexts.

As shown in Figure~\ref{fig:ssl_approach}, the architecture comprises three components: an online encoder, a target encoder, and a predictor. Input sensor windows are first divided into temporal patches via a 3D convolution. A random masking strategy divides patches into visible context and masked targets. The online encoder processes only visible patches, while the target encoder processes the full sequence to provide stable targets. The predictor attempts to reconstruct the representations of masked patches from the visible context. The training objective minimizes the L1 loss between predicted and target representations for masked positions:
\[
\mathcal{L}_{\text{VJEPA}} = \| \mathrm{pred}_{\text{masked}} - \mathrm{target}_{\text{masked}} \|_1
\]

Gradients flow only through the online encoder and predictor; the target encoder is updated as an exponential moving average (EMA) of the online encoder to prevent representational collapse. Like the autoencoder, VJEPA is trained without access to ground-truth health indicators \(x_t\). After training, the online encoder extracts latent representations \(z_t\) for downstream evaluation. Detailed architecture specifications are provided in~\ref{app:jepa_details}.

\section{Experiments}
\label{sec:experiments}

\subsection{Experimental Setup}
\subsubsection{Problem configuration}For all experiments, the models aim to estimate the health indicator (HI) vector $\boldsymbol{x}_t$ at each time step $t$ using the corresponding sensor measurement vector $\boldsymbol{y}_t$ as the primary source of information. Depending on the model typology, additional historical observations may be incorporated to capture temporal dependencies and long‑term system dynamics. For temporal models operating on full sequences, all trajectories are padded with zeros to ensure a uniform sequence length across samples. During training, these padded regions are systematically masked so that they do not contribute to the loss computation or parameter updates, thereby preventing any bias introduced by artificial padding. In addition, all input and output variables are normalized using standard scaling or Min–Max normalization, ensuring stable optimization and comparable feature magnitudes across sensors and HIs.

\subsubsection{Datasets} The full set of degradation trajectories $S$ is divided into three disjoint subsets containing 70\%, 10\%, and 20\% of the sequences, which serve respectively as the training, validation, and test datasets. The training set is used to learn the parameters of the models with trainable components, while the validation set is primarily employed for neural network–based approaches to monitor learning behavior and detect potential overfitting. All methods are evaluated on the same test partition to enable a fair and homogeneous comparison.
To ensure robustness and reduce performance variability due to dataset partitioning, we perform a 5‑fold cross‑validation over the sequence set, ensuring that each degradation trajectory acts once as the test sequence. Final performance metrics are reported as the mean and standard deviation computed across the five folds.

\subsubsection{Hyperparameter tuning}To ensure fair and optimized performance across all models, we perform automated hyperparameter search using the Optuna framework \cite{akiba2019optuna}, a state-of-the-art hyperparameter optimization system based on a define-by-run API and efficient pruning and sampling strategies. Optuna enables the dynamic construction of search spaces and employs advanced algorithms such as Tree-structured Parzen Estimators (TPE) to guide the exploration toward promising regions of the parameter space. During optimization, each model configuration is evaluated using the validation set, and unpromising trials are pruned early to reduce computational cost. The number of trials and search ranges are adapted to the complexity of each model family (e.g., tree-based, feedforward networks, recurrent networks).

\subsubsection{Evaluation criteria}
To evaluate and compare the performance of the models, we employ three complementary criteria: the Symmetric Mean Absolute Percentage Error (SMAPE), the Root Mean Squared Error (RMSE), and the Pearson correlation coefficient (P.Corr.) computed between the ground‑truth health indicators and the predicted values. These metrics were selected because they capture different and equally important aspects of estimation quality. SMAPE provides a scale‑independent measure of relative error, making it suitable for comparing deviations across health indicators with different magnitudes. RMSE emphasizes larger errors due to its quadratic formulation, thereby offering insight into the model’s ability to avoid large deviations that may be critical in prognostics applications. Finally, Pearson correlation measures the linear association between predictions and ground truth, reflecting how well the models capture the underlying trends and degradation dynamics.





\subsection{Experimental Results}

Table 1 reports the performance of tested models for the estimation of the ten HIs from sensor observations. A first observation is that all models achieve good correlation performance across most indicators. This suggests that, despite the underdetermined nature of the treated inverse problem, the simulated sensor signals contain sufficient information to recover the direction of degradation. However, SMAPE show that the overall error is still high, with minimum average value of 14\%. It comes from the cumulative errors for the temporal methods, and from the high noise level for stationary ones.

\begin{table}
\caption{Prediction performance for 10 health indicators across model families.
Results are reported as mean $\pm$ standard deviation over 5-fold cross-validation
using SMAPE, RMSE ($\times10^3$), and Pearson correlation. It should be noted that the mean and std reported in avg column is computed directly on the test sets and is not the average of the 10 columns.}
\resizebox{\textwidth}{!}{
\begin{tabular}{l|l|*{10}{>{\columncolor{white}}c!{\color{lightgray}\vrule}}c}

\toprule
\textbf{Model} & \textbf{Metric} & \textbf{HI1} & \textbf{HI2} & \textbf{HI3} & \textbf{HI4} & \textbf{HI5} & \textbf{HI6} & \textbf{HI7} & \textbf{HI8} & \textbf{HI9} & \textbf{HI10} & \textbf{Avg} \\

\midrule
\multicolumn{13}{c}{\textbf{Steady-state}} \\
\midrule
\multirow{3}{*}{GB} & SMAPE & 0.24 $\pm$ 0.21 & 0.19 $\pm$ 0.20 & 0.30 $\pm$ 0.26 & 0.05 $\pm$ 0.13 & 0.09 $\pm$ 0.14 & 0.18 $\pm$ 0.21 & 0.19 $\pm$ 0.19 & 0.16 $\pm$ 0.18 & 0.24 $\pm$ 0.22 & 0.33 $\pm$ 0.28 & 0.20 $\pm$ 0.09 \\
 & RMSE & 2.0 $\pm$ 0.10 & 1.0 $\pm$ 0.09 & 2.0 $\pm$ 0.08 & 0.3 $\pm$ 0.02 & 1.0 $\pm$ 0.03 & 2.0 $\pm$ 0.10 & 1.0 $\pm$ 0.07 & 1.0 $\pm$ 0.05 & 2.0 $\pm$ 0.03 & 3.0 $\pm$ 0.10 & 1.53 $\pm$ 0.79 \\
 & P.Corr. & 0.93 $\pm$ 0.00 & 0.96 $\pm$ 0.00 & 0.88 $\pm$ 0.01 & 1.00 $\pm$ 0.00 & 0.99 $\pm$ 0.00 & 0.97 $\pm$ 0.00 & 0.96 $\pm$ 0.00 & 0.97 $\pm$ 0.00 & 0.93 $\pm$ 0.00 & 0.87 $\pm$ 0.01 & 0.95 $\pm$ 0.04 \\
\cline{1-13}
\multirow{3}{*}{MLP} & SMAPE & 0.20 $\pm$ 0.25 & 0.15 $\pm$ 0.23 & 0.28 $\pm$ 0.25 & 0.05 $\pm$ 0.15 & 0.07 $\pm$ 0.16 & 0.12 $\pm$ 0.21 & 0.17 $\pm$ 0.22 & 0.12 $\pm$ 0.21 & 0.22 $\pm$ 0.24 & 0.30 $\pm$ 0.30  & 0.17 $\pm$ 0.24\\ 
 & RMSE & 1.24 $\pm$ 0.05 & 0.86 $\pm$ 0.03 & 2.07 $\pm$ 0.08 & 0.19 $\pm$ 0.01 & 0.62 $\pm$ 0.03 & 1.24 $\pm$ 0.05 & 1.11 $\pm$ 0.06 & 0.53 $\pm$ 0.04 & 1.59 $\pm$ 0.04 & 2.19 $\pm$ 0.11 & 1.16 $\pm$ 0.62 \\
 & P.Corr. & 0.97 $\pm$ 0.01 & 0.99 $\pm$ 0.01 & 0.91 $\pm$ 0.01 & 1.00 $\pm$ 0.01 & 1.00 $\pm$ 0.01 & 0.99 $\pm$ 0.01 & 0.98 $\pm$ 0.01 & 1.00 $\pm$ 0.01 & 0.96 $\pm$ 0.01 & 0.91 $\pm$ 0.01 & 0.97 $\pm$ 0.03\\
\midrule
\multicolumn{13}{c}{\textbf{Temporal-based}} \\
\midrule
\multirow{3}{*}{GRU} & SMAPE & 0.12 $\pm$ 0.22 & 0.09 $\pm$ 0.19 & 0.27 $\pm$ 0.24 & 0.04 $\pm$ 0.14 & 0.05 $\pm$ 0.16 & 0.08 $\pm$ 0.19 & 0.16 $\pm$ 0.21 & 0.07 $\pm$ 0.19 & 0.22 $\pm$ 0.23 & 0.28 $\pm$ 0.26 &  0.14 $\pm$ 0.22 \\
 & RMSE & 0.63 $\pm$ 0.05 & 0.48 $\pm$ 0.04 & 2.07 $\pm$ 0.07 & 0.20 $\pm$ 0.01 & 0.39 $\pm$ 0.02 & 0.68 $\pm$ 0.03 & 1.09 $\pm$ 0.05 & 0.31 $\pm$ 0.03 & 1.56 $\pm$ 0.03 & 2.13 $\pm$ 0.11 & 0.95 $\pm$ 0.69 \\
 & P.Corr. & 0.99 $\pm$ 0.01 & 1.00 $\pm$ 0.01 & 0.91 $\pm$ 0.01 & 1.00 $\pm$ 0.01 & 1.00 $\pm$ 0.01 & 1.00 $\pm$ 0.01 & 0.98 $\pm$ 0.01 & 1.00 $\pm$ 0.01 & 0.96 $\pm$ 0.01 & 0.92 $\pm$ 0.01 & 0.97 $\pm$ 0.03 \\ 
\midrule
\multicolumn{13}{c}{\textbf{State-space models}} \\
\midrule
\multirow{3}{*}{K.F.} & SMAPE & 0.10 $\pm$ 0.23 & 0.08 $\pm$ 0.20 & 0.35 $\pm$ 0.37 & 0.03 $\pm$ 0.13 & 0.04 $\pm$ 0.16 & 0.11 $\pm$ 0.28 & 0.18 $\pm$ 0.30 & 0.07 $\pm$ 0.20 & 0.24 $\pm$ 0.27 & 0.26 $\pm$ 0.32 & 0.15 $\pm$ 0.11 \\
 & RMSE &  0.47 $\pm$ 0.01 & 0.42 $\pm$ 0.01 & 2.14 $\pm$ 0.07 & 0.17 $\pm$ 0.00 & 0.36 $\pm$ 0.01 & 0.89 $\pm$ 0.02 & 0.74 $\pm$ 0.02 & 0.35 $\pm$ 0.01 & 1.44 $\pm$ 0.04 & 1.55 $\pm$ 0.05 & 0.85 $\pm$ 0.65 \\
 & P.Corr. & 1.00 $\pm$ 0.00 & 1.00 $\pm$ 0.00 & 0.91 $\pm$ 0.00 & 1.00 $\pm$ 0.00 & 1.00 $\pm$ 0.00 & 1.00 $\pm$ 0.00 & 0.99 $\pm$ 0.00 & 1.00 $\pm$ 0.00 & 0.96 $\pm$ 0.00 & 0.96 $\pm$ 0.00 & 0.98 $\pm$ 0.03 \\
\midrule
\multicolumn{13}{c}{\textbf{Two-stage SSL representational}} \\
\midrule
\multirow{3}{*}{AE} & SMAPE & 0.25 $\pm$ 0.05 & 0.28 $\pm$ 0.05 & 0.24 $\pm$ 0.04 & 0.23 $\pm$ 0.04 & 0.35 $\pm$ 0.12 & 0.38 $\pm$ 0.06 & 0.28 $\pm$ 0.06 & 0.34 $\pm$ 0.09 & 0.26 $\pm$ 0.04 & 0.32 $\pm$ 0.06 & 0.29 $\pm$ 0.05 \\
 & RMSE & 2.53 $\pm$ 0.56 & 2.88 $\pm$ 0.53 & 2.38 $\pm$ 0.36 & 2.33 $\pm$ 0.40 & 3.35 $\pm$ 1.20 & 3.78 $\pm$ 0.67 & 2.83 $\pm$ 0.63 & 3.40 $\pm$ 0.93 & 2.58 $\pm$ 0.33 & 3.18 $\pm$ 0.65 & 2.92 $\pm$ 0.49 \\
 & P.Corr. & 0.88 $\pm$ 0.03 & 0.88 $\pm$ 0.03 & 0.88 $\pm$ 0.03 & 0.88 $\pm$ 0.03 & 0.88 $\pm$ 0.03 & 0.88 $\pm$ 0.03 & 0.88 $\pm$ 0.03 & 0.88 $\pm$ 0.03 & 0.88 $\pm$ 0.03 & 0.88 $\pm$ 0.03 & 0.88 $\pm$ 0.03 \\
\cline{1-13}
\multirow{3}{*}{VJEPA}
& SMAPE & 0.33 $\pm$ 0.01 & 0.32 $\pm$ 0.01 & 0.33 $\pm$ 0.02 & 0.30 $\pm$ 0.03 & 0.19 $\pm$ 0.01 & 0.21 $\pm$ 0.01 & 0.32 $\pm$ 0.03 & 0.31 $\pm$ 0.03 & 0.31 $\pm$ 0.01 & 0.34 $\pm$ 0.01 & 0.30 $\pm$ 0.01 \\
& RMSE & 2.38 $\pm$ 0.15 & 2.44 $\pm$ 0.17 & 2.46 $\pm$ 0.16 & 2.11 $\pm$ 0.37 & 2.23 $\pm$ 0.08 & 2.58 $\pm$ 0.14 & 2.32 $\pm$ 0.08 & 2.29 $\pm$ 0.37 & 2.47 $\pm$ 0.24 & 2.60 $\pm$ 0.21 & 2.40 $\pm$ 0.11 \\
& P.Corr. & 0.89 $\pm$ 0.00 & 0.88 $\pm$ 0.02 & 0.88 $\pm$ 0.01 & 0.91 $\pm$ 0.03 & 0.97 $\pm$ 0.00 & 0.96 $\pm$ 0.00 & 0.90 $\pm$ 0.01 & 0.90 $\pm$ 0.02 & 0.89 $\pm$ 0.02 & 0.87 $\pm$ 0.01 & 0.91 $\pm$ 0.01 \\
\bottomrule
\end{tabular}
}
\end{table}

Another important fact is that performance is consistently worst (across all 3 considered metrics SMAPE, RMSE and correlation) for HI3, HI9 and HI10. This reflects weaker observability of certain degradation factors in the thermodynamic model, and is consistent with prior domain knowledge. Indeed, the Low‑Pressure Turbine is at the end of the air flow in the engine. As a result, it sees all thermodynamic phenomena and is harder to disentangle from others. Moreover, HI3 is the mass flow of the booster compressor, which is thermodynamically inherently intertwined with the fan.

Overall, this weaker performance on structurally less observable HIs highlights an important limitation of purely data-driven inversion. By contrast, the two-stage SSL approaches interestingly exhibit a much more uniform performance across indicators. This suggests that they learned representations capture global structure in the observations, even if they lack the resolution required for precise component-level attribution.

Concerning temporal information, if Steady-state models achieve interesting performances, the non-stationary models are better. As expected, exploiting the \textit{historical context} of the temporal series can help to disambiguate sensor patterns that pertain to similar instantaneous states. Consequently, incorporating temporal structure appears beneficial even when individual observations are informative by themselves\footnote{As for the model mismatch question, knowledge of the exact degradation distribution is of importance. However, we can see here that even knowing perfectly this distribution is not enough to solve the problem.}.

\begin{figure}[H]
    \centering
    \includegraphics[width=1\linewidth]{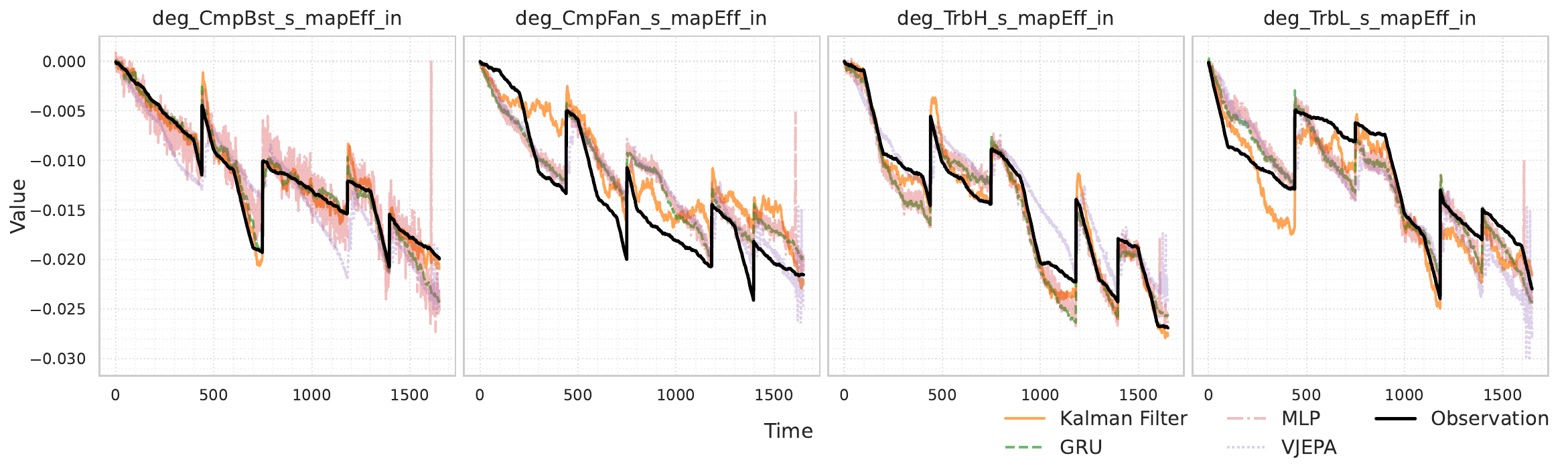}
    \caption{True health indicators versus predictions made by different models for four HI trajectories with varying levels of prediction difficulty.}
    \label{fig:obs_vs_preds}
\end{figure}

Unscented Kalman Filter (UKF) provides the best overall performance on several indicators, using also the time information. However, as previously noted in the global statement, the UKF does not perfectly recover all indicators, leaving room for improvements through hybrid approaches or more accurate degradation dynamics modeling. Indeed, its most important problem is the drift of its estimation through time, and difficulty in presence of maintenance pattern, as shown in Figure \ref{fig:obs_vs_preds}.

In contrast, the two-stage SSL approaches exhibit significantly higher errors, although they still maintain correlations with the ground-truth health indicators. This gap is expected since these methods do not have access to HI labels during representation learning and must rely solely on the structure of the sensor data. Their performance therefore provides an approximate lower bound on how well the health state can be inferred from sensor observations alone without supervision. In consequence, the results indicate that while SSL representations capture a meaningful structure in the data, they remain insufficient for precise component-level estimation without task-specific supervision.

Overall, these results highlight three key insights. First, the turbofan inverse problem remains partially non-observable, with certain indicators systematically harder to estimate. Second, temporal information helps to deal with this difficulty, but none of the algorithms completely solve the different challenges. Third, unsupervised representation learning captures global structure but still falls short of supervised inversion for precise health estimation, suggesting that future work should investigate hybrid strategies combining physics-informed modeling with representation learning.



To better understand the behavior of the different models in predicting the health indicators, we compare their prediction profiles in Figure~\ref{fig:obs_vs_preds}. For clarity and conciseness, four representative indicators are selected based on their varying levels of prediction difficulty. Among them, the Booster Compressor indicator (deg\_CmpBst) exhibits the simplest dynamics and is generally the easiest for the models to estimate. In contrast, the remaining three indicators associated with the Fan, High‑Pressure Turbine, and Low‑Pressure Turbine are difficult HIs as previously discussed, and are difficult to estimate for all models. As shown in the figure, the MLP and GRU models exhibit broadly similar behaviors, both capturing the global trend of the health indicator trajectories. However, the GRU produces smoother estimates with noticeably reduced noise, as it considers temporal dependencies. The Kalman Filter adjusts its predictions according to observed variations (such as changes in degradation speed or maintenance events) and performs well in simpler scenarios. Nevertheless, it struggles when confronted with abrupt or highly nonlinear changes in the system. Finally, VJEPA successfully follows the general evolution of the health indicators, but its predictions remain sensitive to noise in the input signals, and show higher variations for prediction of the trajectories. 

\section{Conclusion}
\label{sec:conclusion}

Estimating turbofan component health from sparse sensor measurements is a fundamentally ill-posed inverse problem. In this work, we introduced a benchmark setting designed to study this challenge under more realistic operational conditions, including heterogeneous degradation dynamics, maintenance interventions, and multiple operating regimes, with hundreds of samples spanning more than 1000 time steps. Within this framework, we compared several modeling paradigms ranging from direct supervised regression and temporal sequence models to state-space filtering and self-supervised representation learning.

Our empirical study highlights three structural properties of the problem. First, while supervised approaches can recover a portion of the health state, performance varies substantially across indicators,
confirming that ambiguity in the inverse mapping is not uniform across the system. Second, the comparison of modeling strategies suggests that temporal structure can help reduce the non-observability of the system. However, no single method is able to accurately recover all HI.
Finally, SSL methods capture global structure in the sensor space but remain insufficient for accurate component-level estimation without supervision. In this sense, these approaches provide a useful reference point for what can be inferred from sensor observations alone, and highlight the gap that task-specific supervision or physics-based constraints must bridge.

Taken together, these results emphasize that turbofan health estimation is a structured inference problem shaped by partial observability, temporal dynamics, and physical constraints. A promising direction for future work is therefore to design hybrid approaches that jointly leverage these three sources of structure. Beyond our methodological comparison, this work also aims to provide the community with a practical experimental framework for studying turbofan health estimation on an industry-oriented dataset.



%
%
%
\bibliographystyle{splncs04}
\bibliography{references}

\newpage
\input{appendix}

\end{document}

%% file: appendix.tex
\appendix
\renewcommand{\thesection}{\appendixname~\Alph{section}}

\section{Notations}
\label{app:notations}

\begin{table}[]
    \centering
    \begin{tabular}{|c|c|}
    \hline
      \textbf{Symbol}   & \textbf{Description}  \\ \hline
       $g$  & Transition Model \\ \hline
       $h$  & Observation Model \\ \hline
       $v,w$ & Gaussian Noise \\ \hline
       $f_\text{proc},f_\text{enc},f_\text{dec}$ & Process, Encoder, Decoder Models \\ \hline
       $x_t$ & HI State Label at timestep $t$ \\ \hline
       $y_t$ & Sensor Measurement \\ \hline
       $z_t$ & Latent Space at timestep $t$ \\ \hline
       $S$ & Trajectory \\ \hline
       $\text{OC}$ & Operating Condition \\ \hline
       $\tau$ & Forecasting Horizon \\ \hline
       $\phi$ & Downstream Task Block Predictor \\ \hline
       $m_t$ & Maintenance action at timestep $t$ \\ \hline
    \end{tabular}
    \caption{Proposed Notation (to refactor in text).}
    \label{tab:naming_convention}
\end{table}

\section{Simulation parameters}
\label{app:sim_parameters}
To generate the degradation trajectories for each health indicator, a set of intervals with minimum and maximum values are considered. Each indicator value represents a deviation/delta from the nominal value for efficiency or corrected Mass flow (Wc) in a module, applied as a scaling factor to the module map. Each indicator has a specific bound, expressing the possible degradation or variation for each module. These bounds are shown in Table \ref{tab:hi_bounds}.

\begin{table}[ht]
    \centering
    \caption{Health Indicators (HIs): degradation factors per module with their physical bounds. Each indicator represents a subsystem state used to generate observations.}
    \label{tab:hi_bounds}
    \begin{tabular}{l l l c c}
        \toprule
        \textbf{State Label} & \textbf{Module} & \textbf{Quantity} & \textbf{Min} & \textbf{Max} \\
        \midrule
        \texttt{deg\_CmpFan\_s\_mapEff\_in} & Fan & Efficiency & -0.05 & 0.00 \\
        \texttt{deg\_CmpFan\_s\_mapWc\_in}  & Fan & Mass Flow  & -0.05 & 0.03 \\
        \midrule
        \texttt{deg\_CmpBst\_s\_mapEff\_in} & Booster Compressor & Efficiency & -0.05 & 0.00 \\
        \texttt{deg\_CmpBst\_s\_mapWc\_in}  & Booster Compressor & Mass Flow  & -0.05 & 0.03 \\
        \midrule
        \texttt{deg\_CmpH\_s\_mapEff\_in}   & High-Pressure Compressor & Efficiency & -0.05 & 0.00 \\
        \texttt{deg\_CmpH\_s\_mapWc\_in}    & High-Pressure Compressor & Mass Flow  & -0.05 & 0.03 \\
        \midrule
        \texttt{deg\_TrbH\_s\_mapEff\_in}   & High-Pressure Turbine & Efficiency & -0.05 & 0.00 \\
        \texttt{deg\_TrbH\_s\_mapWc\_in}    & High-Pressure Turbine & Mass Flow  & -0.05 & 0.05 \\
        \midrule
        \texttt{deg\_TrbL\_s\_mapEff\_in}   & Low-Pressure Turbine & Efficiency & -0.05 & 0.00 \\
        \texttt{deg\_TrbL\_s\_mapWc\_in}    & Low-Pressure Turbine & Mass Flow  & -0.05 & 0.05 \\
        \bottomrule
    \end{tabular}
\end{table}

For example:
\begin{itemize}
    \item \texttt{deg\_CmpFan\_s\_mapEff\_in} can take values from -0.05 (max degradation of 5 \%pt) to 0.0 (nominal);
    \item \texttt{deg\_CmpFan\_s\_mapWc\_in} goes from -0.05 to +0.03, etc.
\end{itemize}

\section{Descriptive analysis}
\label{app:descriptive_analysis}
\begin{figure}
    \centering
    \includegraphics[width=.8\linewidth]{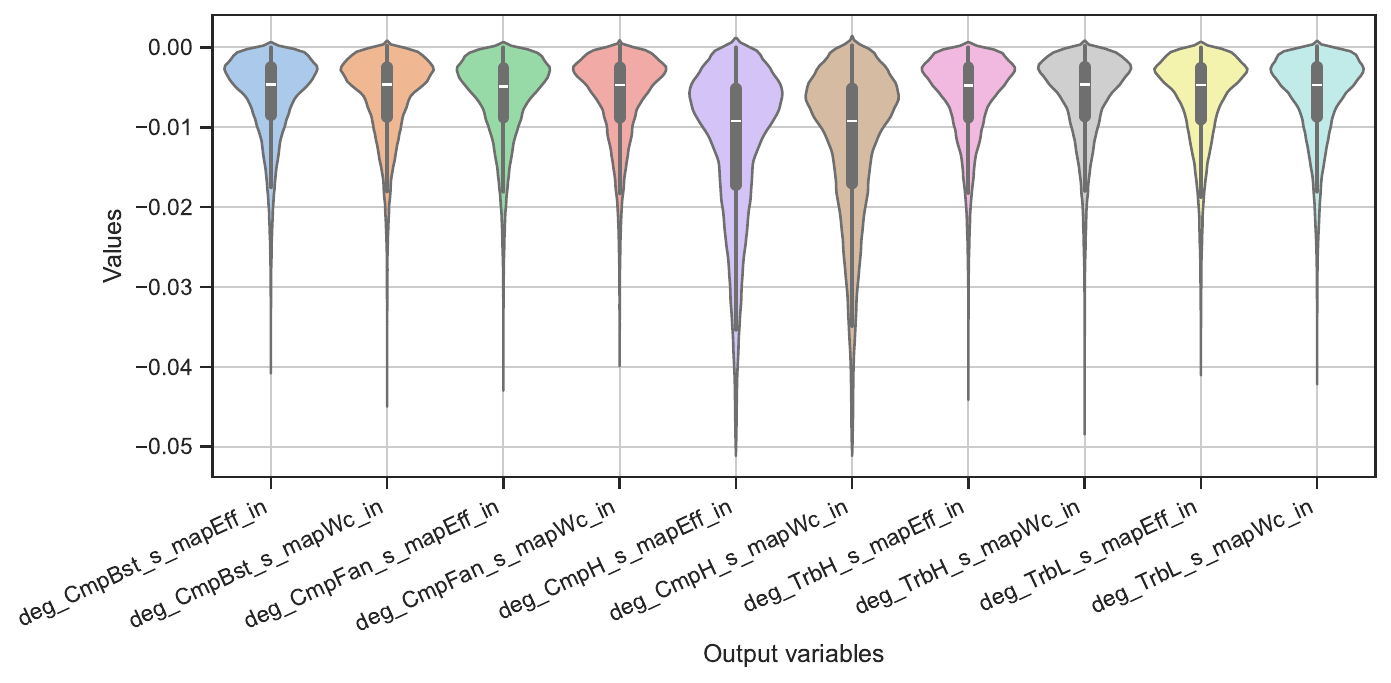}
    \caption{health indicators distribution}
    \label{fig:hi_dist}
\end{figure}

The distribution of the health indicators is presented in Figure~\ref{fig:hi_dist}. Most indicators exhibit similar statistical profiles, reflecting consistent behavior across the degradation trajectories. However, two indicators—both associated with the high‑pressure compressor (HPC), specifically its efficiency and mass flow—display noticeably broader and more dispersed distributions. This increased variability is expected, as the HPC operates under significantly higher thermal and mechanical stresses, making its behavior inherently more volatile. Consequently, these indicators are likely to present greater prediction difficulty for the models, given their higher sensitivity to operating conditions and measurement fluctuations.

\begin{figure}
    \centering
    \includegraphics[width=1\linewidth]{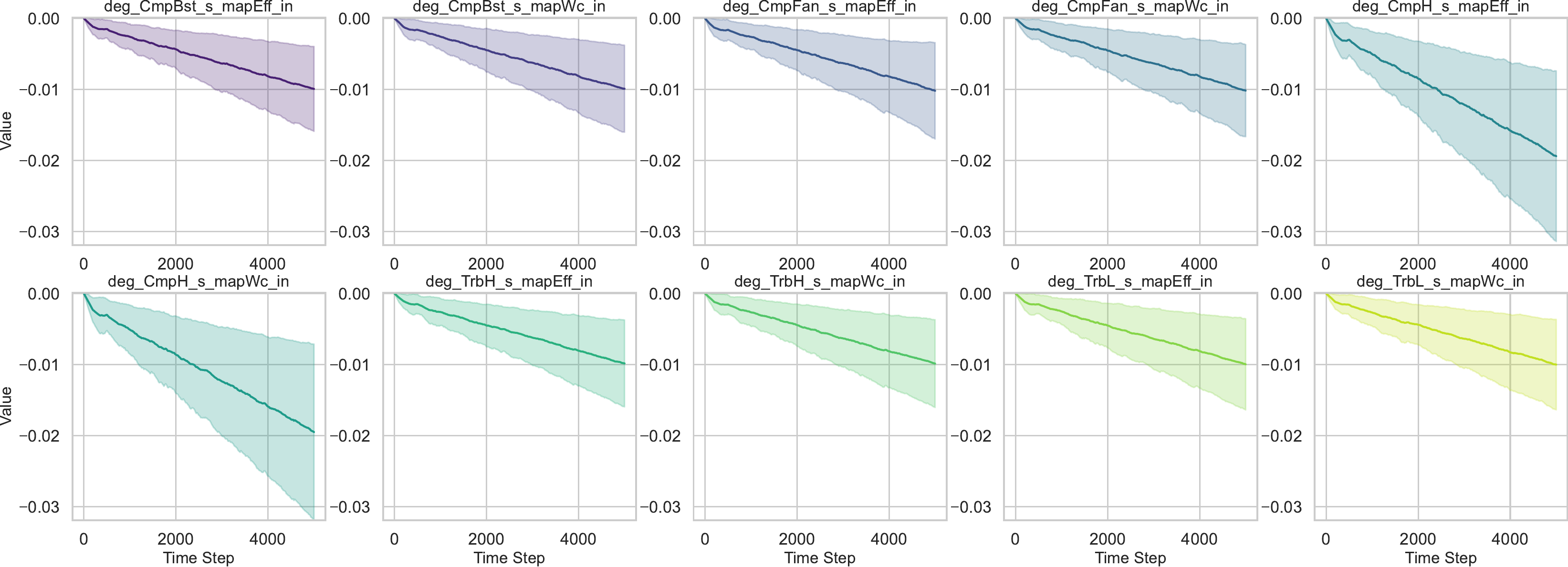}
    \caption{Health indicators (degradation trajectories) mean and standard deviation over all the sequences}
    \label{fig:degradation_dist}
\end{figure}

To characterize the degradation behavior, the ten health indicators were aggregated and visualized using mean trajectories with associated standard deviation bands in Figure \ref{fig:degradation_dist}. Because the trajectories differed in length, each indicator was first interpolated onto a common time grid to allow consistent temporal alignment. The resulting plots reveal smooth and strongly persistent degradation trends, with mean values exhibiting monotonic or quasi‑monotonic evolution over time. The width of the standard deviation bands provides insight into inter‑trajectory variability, where narrow regions indicate consistent degradation dynamics across engines, whereas wider regions correspond to indicators more sensitive to operating conditions, sensor noise, or maintenance interventions. 


\begin{figure}
    \centering
    \includegraphics[width=1\linewidth]{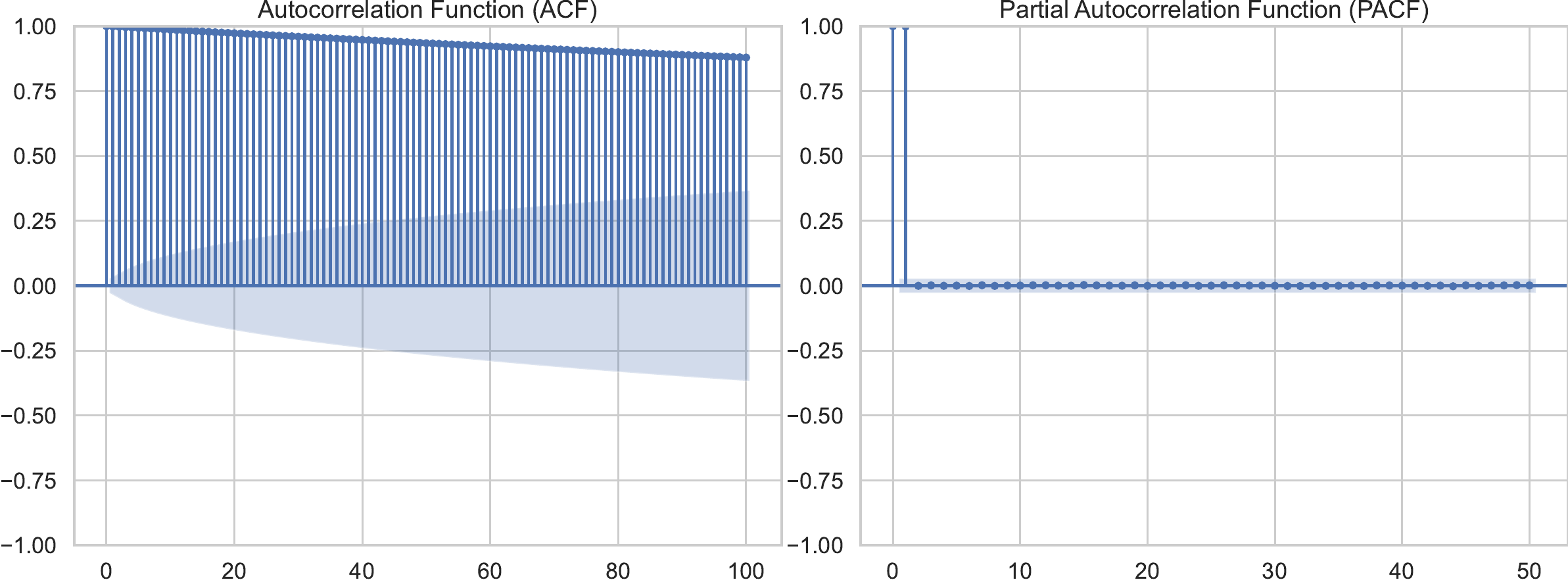}
    \caption{Auto-correlation (ACF) and partial autocorrelation (PACF). The x-axis represents the lag and y-axis represents the correlation value (-1,1)}
    \label{fig:acf}
\end{figure}

The auto-correlation (ACF) and Partial auto-correlation functions (PACF) are computed for one sequence and the Booster Compressor efficiency indicator (\textit{deg\_CmpBst\_s\_mapEff\_in}) and the results are shown in Figure \ref{fig:acf}. ACF of the degradation trajectories shows extremely high autocorrelation values across all lags, with a very slow decay, indicating strong temporal persistence and non‑stationarity. The PACF exhibits a dominant spike at lag 1 followed by values within the confidence bands, which is characteristic of near‑unit‑root or random‑walk‑like processes. This implies that health indicators evolve smoothly over time, with each state being highly dependent on the immediately preceding one. Such behavior is typical in physical degradation processes where wear accumulates gradually rather than abruptly.

\section{JEPA Architecture Details}
\label{app:jepa_details}

\subsection{Input Processing}
Each input sample is a window of sensor measurements \(y_t \in \mathbb{R}^{T \times C}\) with temporal window size \(T = 50\) and \(C\) representing the total number of sensor channels (product of sensors and operating contexts). A 3D patch embedding with kernel size \((16,1,1)\) and stride \((16,1,1)\) divides the window into non-overlapping temporal patches, each covering 16 consecutive time steps, producing a sequence of \(N = \lceil 50/16 \rceil = 4\) patches. Each patch is projected to a \(D\)-dimensional embedding, and a learned positional embedding is added to retain temporal order.

\subsection{Encoder Architectures}
Both online and target encoders are Transformers with depth 32, embedding dimension \(D=36\), and 6 attention heads. They process only visible patches (online) or full sequences (target) using standard Transformer encoder layers with GELU activation.

\subsection{Predictor}
The predictor is a shallower Transformer (depth 32) that takes the online encoder's output and the indices of masked patches. It constructs a full sequence by inserting a learnable mask token at each masked position, processes this sequence, and outputs predictions only for the masked positions.

\subsection{Training Details}
Random masking uses a ratio of 0.75, selecting masked patches uniformly at random. The target encoder is updated via exponential moving average: \(\theta_{\text{target}} \leftarrow \tau \theta_{\text{target}} + (1 - \tau) \theta_{\text{online}}\) with \(\tau = 0.99\). The model is trained with the L1 loss between predicted and target representations for masked patches only.

\section{Latent Visualisation}

\begin{figure}
    \centering
    \includegraphics[width=\textwidth]{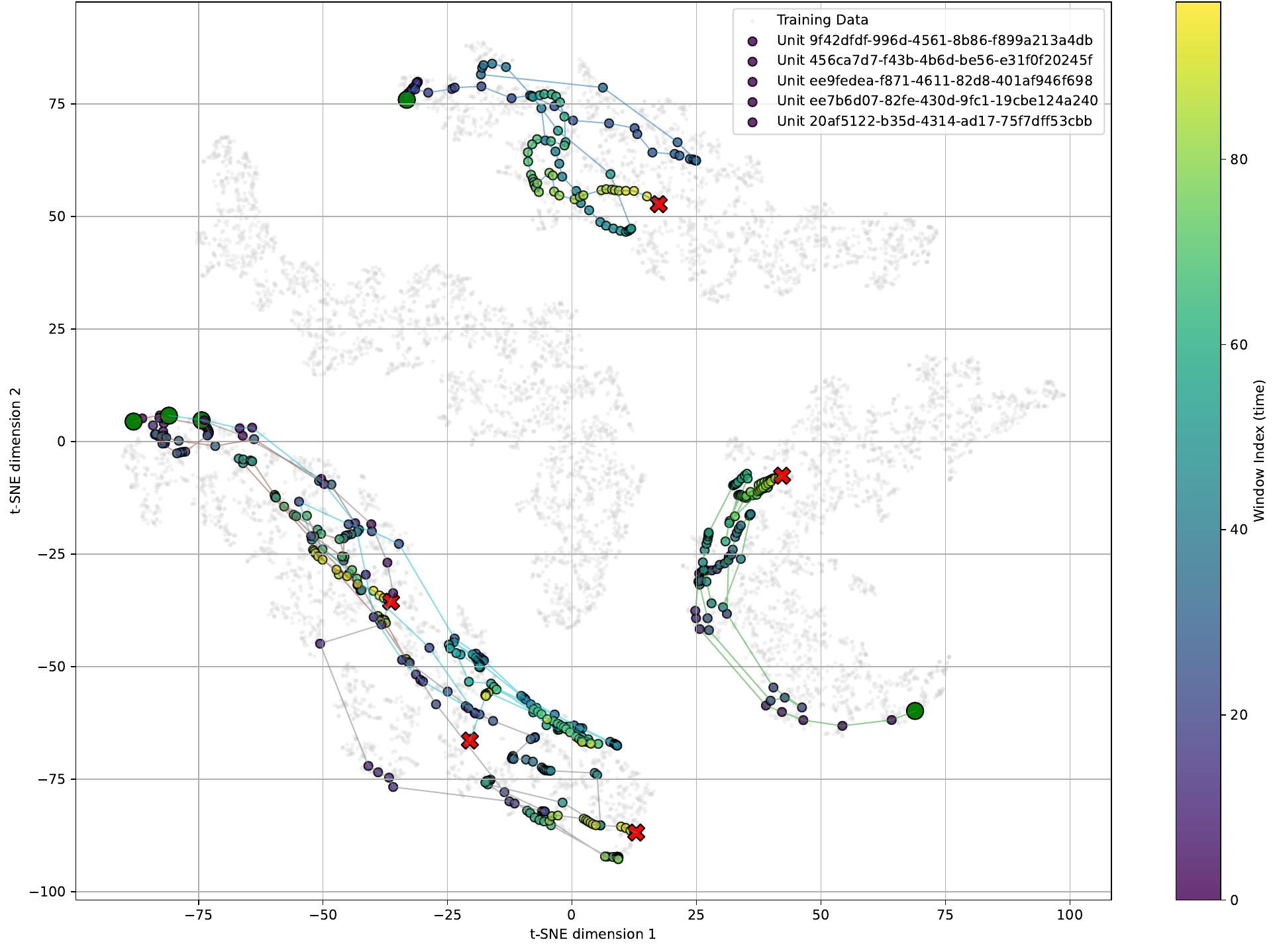}
    \caption{t-SNE plot of few trajectories from the VJEPA architecture. The four clusters correspond to trajectories grouped by the amount of maintenance performed.}
    \label{fig:VJEPA2_multiple_trajectories}
\end{figure}

\section{Enabling further experimental approaches}

\begin{figure*}[ht]
    \centering
    \centerline{\includegraphics[width=\textwidth]{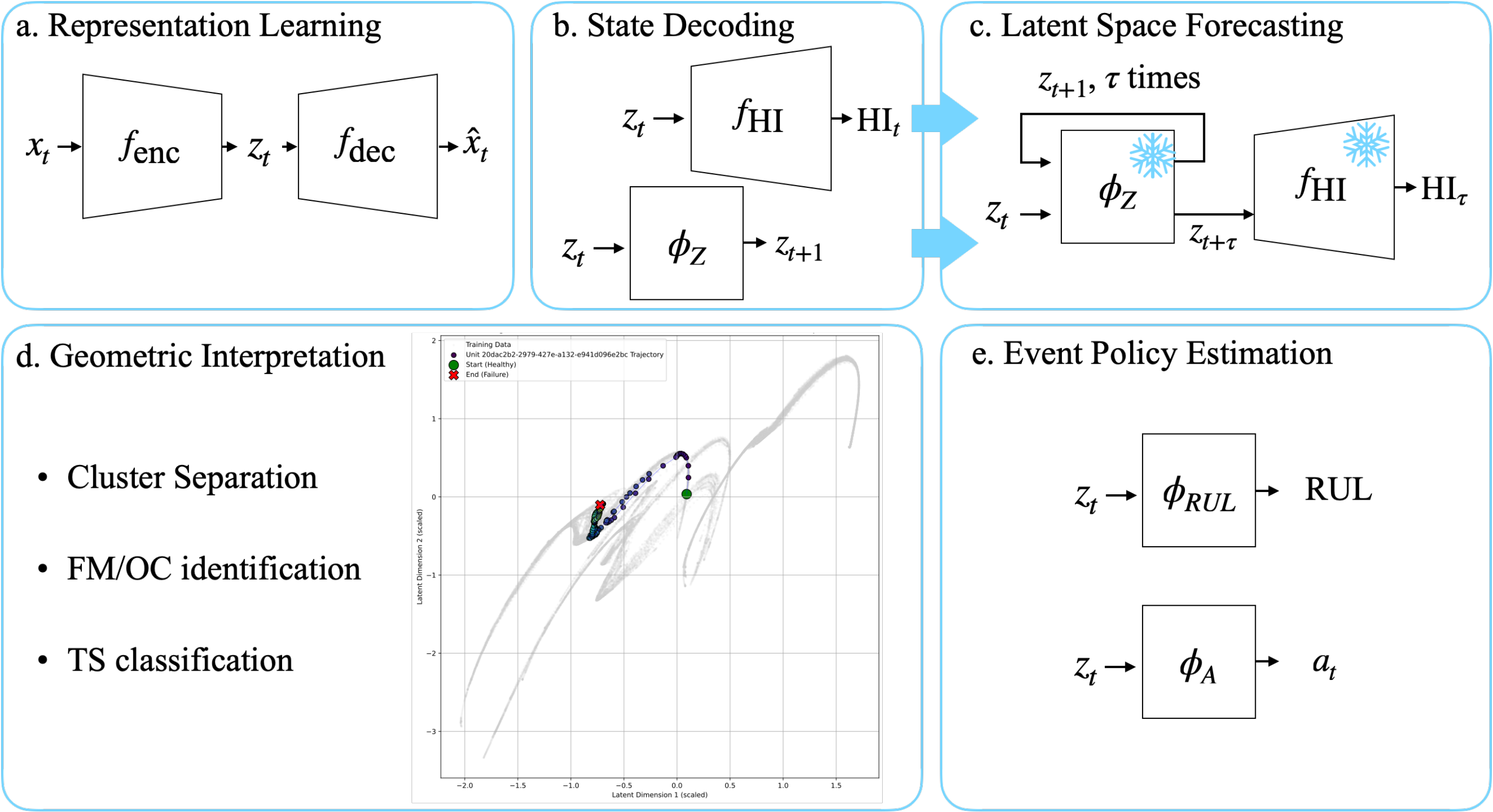}}

    \caption{(Proposal) \textbf{Multitask Experiment Setup Overview}. \textbf{a) Learning:} A model learns a representation from sensor observation which is evaluated on a series of downstream tasks. \textbf{b) Decoding:} The latent $z$ is used to train a regression head to estimate the HI labels, and another block $\phi_z$ is trained for next state prediction $z_{t+1}=\phi_z(z_t)$. \textbf{c) Forecasting:} the frozen blocks are used for multi-state forecasting over $\tau$ timesteps from $n$ randomly sampled initial states $z_t$. We compute the $R^2$ score of the estimated HIs through time. \textbf{d) Geometric Interpretation:} Through clustering methods, we investigate if the learned representation reflects the underlying geometry in terms of cluster separation of signature degradation modes and maintenance history. \textbf{e) Prescription:} we evaluate the model capabilities in RUL estimation as well as the horizon of maintenance actions. 
    }
    
\end{figure*}

\subsubsection{HI Forecasting}
We investigate if the learned representations $z$ are stable for HI forecasting on different horizons over $\tau$ steps. We start from $n$ random indices in the test data and compute the forecasting $R^2$ score. Ideally, the representation remains predictive over long horizons, leading to a high $R^2$ score.

We train a predictor $\phi_z$ to model latent dynamics: $\phi(z_t) \mapsto z_{t+1}$. At test time, we start from the true latent state $z_t = f_{\text{enc}}(x_t)$ and perform an autoregressive rollout by applying $\phi$ recursively for 15 steps (each step representing one second after discretization) without access to further ground-truth data: $z_{t+k} = \phi_z(z_{t+k-1})$ for $k=1,\dots,15$. The final prediction is reconstructed as $\hat{x}_{t+15} = f{\text{dec}}(z_{t+15})$. We compare the different methods together and evaluate the influence of different window sizes for backbone architectures, hyperparameters, and a vanilla autoencoder baseline. We report the $\text{R}^2$ score averaged over $n$ randomly sampled starting times $t$.

\begin{equation}
    \hat{x}_{t+\tau} = f_{\text{dec}}(z_{t+\tau})
\end{equation}

\subsubsection{Event anticipation}
Similar to RUL prediction, but extended to action prediction where we estimate the amount of remaining timesteps until a action (e.g maintenance) must be performed. It acts as a proxy for subphase degradation estimation while dissociating it from the general RUL prediction (overall degradation). We note respectively $a_t = \phi_\text{A}(z_t)$ and $\text{RUL} = \phi_\text{RUL}(z_t)$ the blocks for action and RUL estimation.

\subsubsection{Profile Clustering}
A visual/empirical analysis if the learned model or representations can be interpretable, if we can obtain insights on the functional regimes of the dataset: clear separation of failure modes, isolation of operating conditions, signature degradations. This can be linked to OOD detection.

\subsubsection{OOD Detection}
We explicitly test the model's ability to detect OOD samples through simulating a new, unseen fault mode, and running the engine under an extreme operating condition not present in the training data. The results are evaluated via reconstruction error and time series classification of the trajectories. 
